\documentclass[runningheads]{llncs}

\usepackage{graphicx}
\usepackage{comment}
\usepackage{amsmath,amssymb}
\usepackage{color}
\usepackage{url}
\usepackage{hyperref}
\usepackage[capitalize]{cleveref}
\usepackage{nicefrac}
\usepackage{todonotes}
\usepackage{subcaption}
\usepackage{xspace}
\usepackage{placeins}
\usepackage{booktabs}
\usepackage[T1]{fontenc}
\usepackage{microtype}

\crefname{section}{Sec.}{Secs.}%

\graphicspath{ {./images/} }

\newcommand{\Rt}{\ensuremath{\mathcal{R}_t}\xspace}

\newif\ifreview
\reviewfalse

\ifreview
	\usepackage{lineno}

	\linenumbers
\fi

\begin{document}

\def\SubNumber{035}

\def\GCPRTrack{Special Track: Pattern recognition in the life and natural sciences}

\title{Modeling COVID-19 Dynamics in German States Using Physics-Informed Neural Networks}

\ifreview
\else
	\titlerunning{Modeling COVID-19 Dynamics in German States}

	\author{Phillip Rothenbeck\orcidID{0009-0002-7864-9013} \and
    Sai Karthikeya Vemuri\orcidID{0009-0003-6272-8603} \and
	Niklas Penzel\orcidID{0000-0001-8002-4130} \and
	Joachim Denzler\orcidID{0000-0002-3193-3300}}
	
	\authorrunning{P. Rothenbeck et al.}
	
	\institute{Computer Vision Group, Friedrich Schiller University, Inselplatz, Jena, 07743, Thuringia, Germany
	\email{\{firstname.lastname\}@uni-jena.de}}
\fi

\maketitle              %

\begin{abstract}%

The COVID-19 pandemic has highlighted the need for quantitative modeling and analysis to understand real-world disease dynamics.
In particular, post hoc analyses using compartmental models offer valuable insights into the effectiveness of public health interventions, such as vaccination strategies and containment policies. 
However, such compartmental models like SIR (Susceptible-Infectious-Recovered) often face limitations in directly incorporating noisy observational data. 
In this work, we employ Physics-Informed Neural Networks (PINNs) to solve the inverse problem of the SIR model using infection data from the Robert Koch Institute (RKI). 
Our main contribution is a fine-grained, spatio-temporal analysis of COVID-19 dynamics across all German federal states over a three-year period. 
We estimate state-specific transmission and recovery parameters and time-varying reproduction number (\Rt) to track the pandemic progression. 
The results highlight strong variations in transmission behavior across regions, revealing correlations with vaccination uptake and temporal patterns associated with major pandemic phases. 
Our findings demonstrate the utility of PINNs in localized, long-term epidemiological modeling.

\end{abstract}
\section{Introduction}

\begin{figure}[t]
    \centering
    \begin{subfigure}{0.44\textwidth}
        \includegraphics[width=\textwidth, trim=0cm 0cm 0cm 0.89cm, clip]{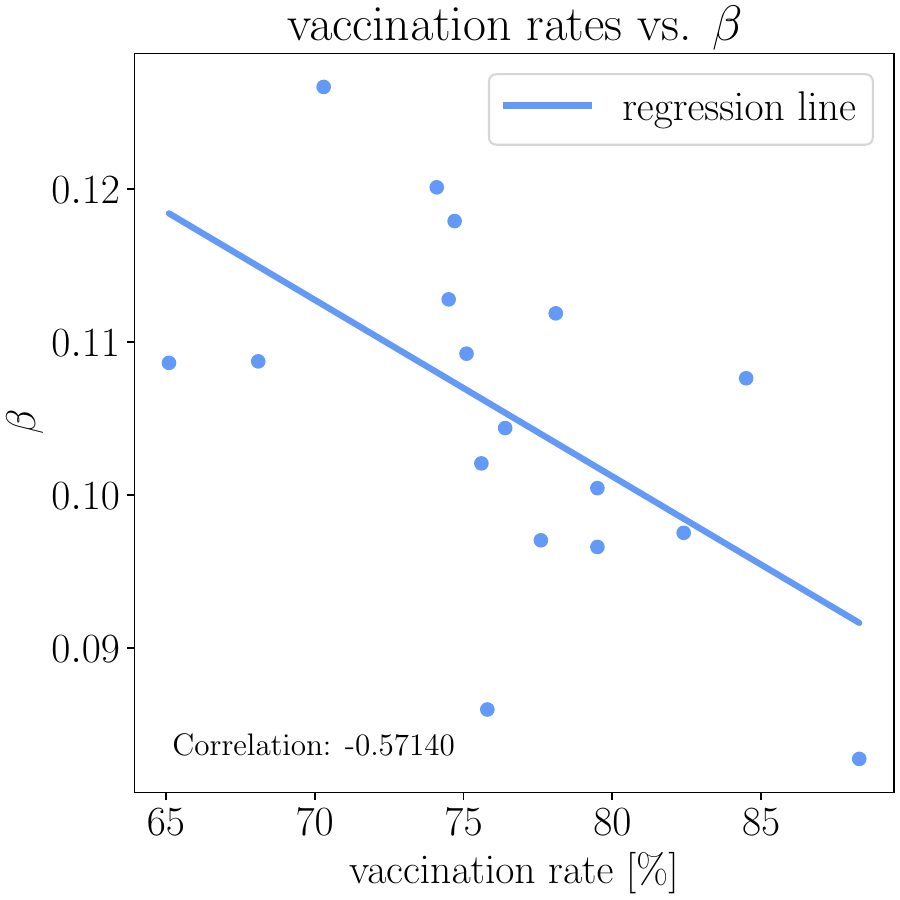}
    \end{subfigure}
    \begin{subfigure}{0.44\textwidth}
        \includegraphics[width=\textwidth, trim=0cm 0cm 0cm 0.89cm, clip]{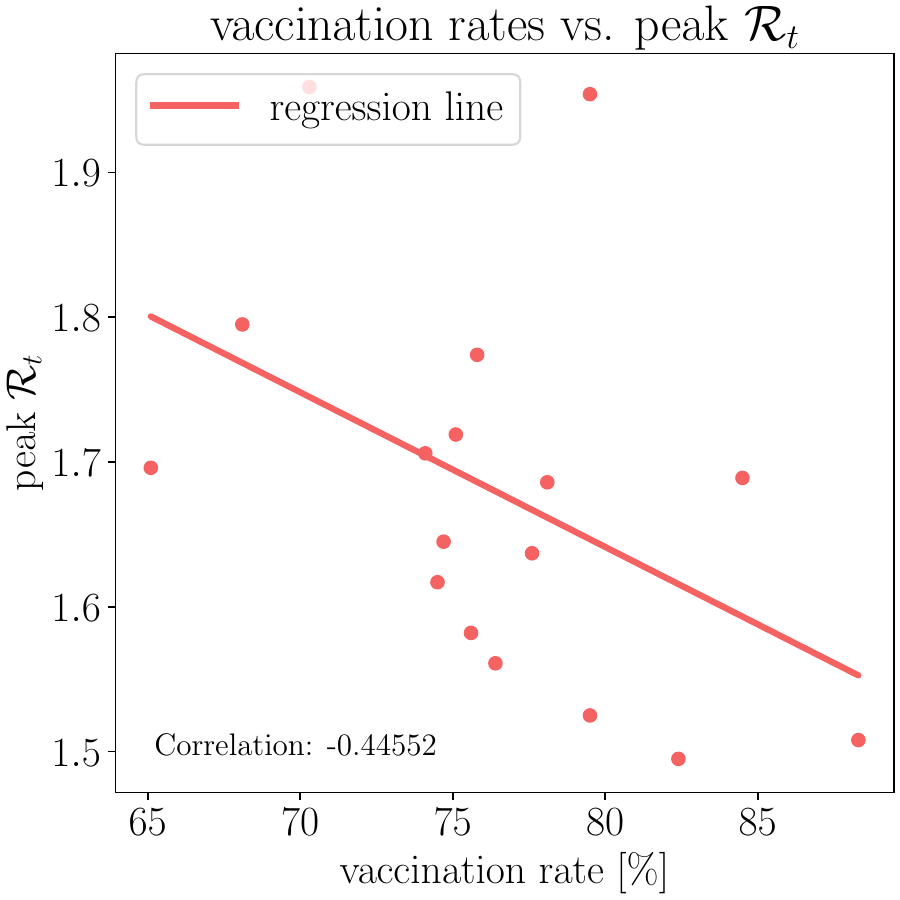}
    \end{subfigure}
    \vspace{-0.2cm}
    \caption{
    Higher vaccination coverage coincides with lower pandemic effects.
    (Left): The correlation between the vaccination rate and the corresponding mean transmission rate $\beta$ for each federal state. 
    (Right): The correlation between the vaccination rate and the peak $\Rt$ value for each state.
    }
    \label{fig:beta_correlation}
\end{figure}

The COVID-19 pandemic posed a global public health crisis, revealing stark regional differences in how outbreaks unfolded and were managed~\cite{Bhatkar2023-fy,GAGNON2023101258,RKIa}. 
In Germany, federal states exhibited diverse trajectories due to variations in policies, behavior, and healthcare infrastructure. 
Capturing these state-level dynamics is essential for evaluating public health responses and informing future interventions.

To simulate disease progression, compartmental models, such as the Susceptible-Infectious-Recovered (SIR) framework~\cite{Kermack1927}, remain foundational in epidemiology.
Such models divide populations into discrete health compartments. 
However, they often assume static parameters and are not designed to incorporate noisy, real-world data. 
Physics-Informed Neural Networks (PINNs)~\cite{Raissi2019} address this limitation by embedding differential equations into network training, enabling robust estimation of latent parameters directly from observed data.

To extract deeper insights from pandemic data, researchers typically either expand model complexity, e.g., by adding compartments for vaccination or hospitalization~\cite{Berkhahn2022,han2024approaching}, or increase spatio-temporal resolution by analyzing local regions and time-varying effects. 
We pursue the latter using a PINN-based framework to incorporate real-world observations.

Specifically, our main contribution is a spatio-temporal analysis of COVID-19 dynamics across all 16 German federal states over a period of 1,200 days. 
Using publicly available infection data from the Robert Koch Institute (RKI) \cite{RKI,RKIa}, we solve the inverse problem of estimating the transmission ($\beta$) and recovery ($\alpha$) parameters of the SIR model using PINNs for each individual state. 
Building on these results, we estimate the time-dependent reproduction number $\Rt$ \cite {Millevoi2023} for each state, providing temporal insights into how transmission evolved across pandemic phases, such as the emergence of variants or vaccination campaigns.

We find that variations in state-level vaccination rates correlate with both estimated transmission rates $\beta$ and observed peak reproduction numbers $\Rt$ (see \cref{fig:beta_correlation}). 
Such regional differences suggest that local interventions had measurable impacts on transmission dynamics, underscoring the value of localized analyses.
 
To substantiate these findings, the following sections will first outline our PINN-based methodology for estimating pandemic parameters from observational data. 
We then present a detailed spatio-temporal analysis of the results for all 16 German federal states, before discussing how these dynamics correlate with real-world public health measures.

\section{Related Work}
\subsection{COVID-19 Dynamics Analyses - Classical Methods}
In \cite{schaback2020covid19}, the authors discuss various comparably simple numerical and mathematical SIR models to model pandemic dynamics specifically for COVID-19.
Similarly, \cite{rahman2021review} surveys various studies introducing COVID-19 modeling strategies.
They focus on three countries (China, the UK, and Australia), developing a framework for regional areas.
In \cite{rahman2023modelling}, the authors concentrate on New South Wales and investigate control strategies in rural districts as well as densely populated districts.
Here, a more complex SEIR-X model is employed.
Closely related to our work is \cite{baerwolff2021modeling}, where the COVID-19 pandemic in European countries is studied using an SIR model.
They estimate the transmission rate $\beta$ on data measured in Germany using the classical dampened Gauss-Newton method.

In contrast to these works, we split our analysis at the level of the German federal states.
Further, we employ the framework of Physics-Informed Neural Networks (PINNs) \cite{Raissi2019} to seamlessly integrate real-world observations.
Additionally, inspired by \cite{Millevoi2023}, we study a longer period of the pandemic using a time-dependent reduced SIR model to estimate the reproduction number \Rt.
\subsection{PINNs for Epidemiology}
Several studies have applied Physics-Informed Neural Networks (PINNs) \cite{Raissi2019} to model COVID-19 and other infectious diseases. 
For instance, Disease-Informed Neural Networks (DINN) were introduced in \cite{Shaier2021}, demonstrating the ability of PINNs to forecast epidemic trajectories. 
Their approach was validated on 11 diseases using the SIDR (Susceptible-Infectious-Dead-Recovered) model to estimate relevant parameters effectively.
A more complex SVIHR (Susceptible-Vaccinated-Infectious-Hospitalized-Removed) model was employed in \cite{Berkhahn2022} to analyze COVID-19 dynamics in Germany, covering data up to the end of 2021. 
Their study compared PINNs with the non-standard finite differences (NSFD) method and showed that PINNs effectively adapted to changing vaccination rates and emerging variants. 
Similarly, in \cite{han2024approaching}, the authors studied the beginning of the pandemic in Germany using SIR models and a more complex SAIRD model including asymptomatic and dead compartments.
In contrast, our work extends the SIR analysis over a longer time period and across all German federal states to gain insights into local differences.
Regarding the progress of the pandemic, in \cite{Olumoyin2021}, the authors estimated time-dependent transmission rates in an asymptomatic-SIR model. 
Their PINN approach leveraged cumulative infection and recovery data to assess the impact of mitigation measures. It was applied to COVID-19 data from Italy, South Korea, the UK, and the US, highlighting the role of vaccinations and non-pharmaceutical interventions.
Lastly, \cite{Millevoi2023} explored time-dependent transmission changes using the reproduction number $\Rt$, introducing a reduced-split PINN approach that alternated between data fitting and minimizing the residuals of differential equations. 
Their method, tested on synthetic and real-world data from early outbreaks in Italy, improved accuracy and training efficiency.
In contrast to these studies, our work estimates key epidemiological parameters ($\alpha$, $\beta$) and the time-dependent reproduction number ($\Rt$) at the state level in Germany. 
By analyzing the period from March 2020 to June 2023, we provide a more comprehensive understanding of how COVID-19 evolved across different German states over three years, offering insights into regional variations in transmission and recovery rates.
\section{Theoretical Foundation}

\subsection{PINNs for Inverse Problems}
In inverse problems, the goal is to infer unknown system parameters (e.g., coefficients, boundary/initial conditions in PDEs) from partial or noisy observations of the system's behavior~\cite{bookinvprob}.
Physics-Informed Neural Networks (PINNs)~\cite{Raissi2019} address these by embedding the governing partial differential equations (PDEs) directly into the neural network's training process.

\subsubsection{Mathematical Formalization:}
Consider a system described by a PDE over a spatial domain $\Omega$ and time $t \in [0, T]$:
\begin{equation}\label{eq:pde}
    \mathcal{F}(u(\mathbf{x}, t), \nabla u(\mathbf{x}, t), \nabla^2u(\mathbf{x},t), ...| \mathbf{\lambda}) = 0, \quad \text{in} \ \Omega \times [0, T],
\end{equation}
where $u(\mathbf{x}, t)$ is the variable of interest (e.g., number of infected people), $\mathcal{F}$ is a differential operator, and $\mathbf{\lambda}$ are unknown parameters to be inferred. 
Given observations $\{(\mathbf{x}_i, t_i, u_i)\}$ of the system, the inverse problem is to estimate $ \mathbf{\lambda} $ such that $ u(\mathbf{x}, t) $ satisfies both the physical laws (\cref{eq:pde}) and the observed data.

PINNs approximate $u$ with a neural network $u_\theta(\mathbf{x}, t)$ (where $\theta$ are the trainable parameters) and learn $\mathbf{\lambda}$ by minimizing a composite loss function~\cite{Raissi2019,soilsai}.

\subsubsection{Loss Function: }
The total loss function for an inverse problem combines data fidelity and physics consistency and is defined as $\mathcal{L}(\theta, \mathbf{\lambda}) = \mathcal{L}_{\text{data}}(\theta) +  \mathcal{L}_{\text{physics}}(\theta, \mathbf{\lambda})$.
The \emph{data loss} $\mathcal{L}_\text{data}$ ensures that $u_\theta$ matches observations:
\begin{equation}\label{eq:loss-data}
    \mathcal{L}_{\text{data}}(\theta) = \frac{1}{N} \sum_{i=1}^{N} \left| u_\theta(\mathbf{x}_i, t_i) - u_i \right|^2.
\end{equation}
The \emph{physics loss} $\mathcal{L}_\text{physics}$ penalizes deviations from the PDE (\cref{eq:pde}):
\begin{equation}\label{eq:loss-physics}
    \mathcal{L}_{\text{physics}}(\theta, \mathbf{\lambda}) = \frac{1}{M} \sum_{j=1}^{M} \left| \mathcal{F}(u_\theta(\mathbf{x}, t), \nabla u_\theta(\mathbf{x}, t), \nabla^2u_\theta(\mathbf{x},t), ...| \mathbf{\lambda}) \right|^2.
\end{equation}
The unknown PDE parameters $\mathbf{\lambda}$ are treated as trainable variables alongside the neural network weights $\theta$, optimized by minimizing $\mathcal{L}(\theta, \mathbf{\lambda})$.
Additional balancing of the loss terms using scalar hyperparameters is possible. 

\subsection{Compartmental Models for Epidemiology and Inverse Problem}
\label{sec:sir-models}
Compartmental models are the foundation of mathematical epidemiology~\cite{Hethcote1989,Kermack1927,Mwalili2020} to study the spread of diseases. 
The SIR model~\cite{Kermack1927} partitions a population of size N into three distinct compartments:
\begin{itemize}
    \item \textbf{Susceptible} (S) individuals at risk of infection.
    \item \textbf{Infected} (I) individuals capable of transmitting the disease.
    \item \textbf{Removed} (R) individuals recovered with immunity or deceased.
\end{itemize}

\noindent The evolution of these compartments over time is governed by the following system of ordinary differential equations (ODEs):
\begin{align}\label{eq:sir}
    \frac{dS}{dt} = -\beta\frac{S I}{N} &, & \frac{dI}{dt} = \beta\frac{S I}{N} - \alpha I &, &\frac{dR}{dt} = \alpha I, 
\end{align}
where \textbf{$\beta$} is the \textbf{transmission rate} and \textbf{$\alpha$} is the \textbf{recovery rate}.

Real-world infectious diseases exhibit dynamic behavior due to changing intervention strategies, population immunity, and viral mutations. 
A key metric to measure this is the effective reproduction number $\Rt$~\cite{rahman2021review}, quantifying the average secondary infections from one infected individual at time $t$:
\begin{align} \label{eq:rt}
\Rt = \frac{\beta(t)}{\alpha(t)} \cdot \frac{S(t)}{N}.
\end{align}
Specifically, an outbreak is expanding if $\Rt>1$ and declining if $\Rt<1$~\cite{Bhatkar2023-fy,Millevoi2023}.

To model time-varying dynamics, Millevoi et al.~\cite{Millevoi2023} reformulated SIR using a rescaled time-dependent formulation, assuming a constant recovery rate $\alpha$. 
Let $t_s$ be a normalized time variable $t_s = \nicefrac{(t - t_0)}{(t_f - t_0)}$, for an interval $t \in [t_0, t_f]$.
Then the infected compartment $I(t)$ is scaled by a constant $c$ as $I(t) = c \cdot I_s(t_s)$, where the dynamics of the scaled infected compartment $I_s$ are given by:
\begin{align} \label{eq:rt-ode}
\frac{dI_s}{dt_s} = \alpha (t_f - t_0)(\Rt - 1) I_s(t_s). 
\end{align}
\section{Methodology}
\label{sec:method}
Here we specify our concrete methodology for solving the inverse problems connected to the compartmental models in \cref{sec:sir-models}.
First, we detail the time-independent analysis of the SIR rates before we discuss how to determine \Rt. 

\subsection{Time-Independent Parameter ($\alpha$ and $\beta$) Identification}
\label{sec:method1}
To identify $\alpha$ and $\beta$, we train a PINN to fit both the SIR model and observed data, as introduced in~\cite{Shaier2021}. 
Hence, \cref{eq:loss-data} becomes the mean squared error (MSE) between predictions and observed data over a period of time $T$:
\begin{equation} 
\mathcal{L}_{data} = \frac{1}{T} \sum_{t=1}^{T} \Big(|\hat{S}^{(t)} - S^{(t)}|^2 + |\hat{I}^{(t)} - I^{(t)}|^2 + |\hat{R}^{(t)} - R^{(t)}|^2 \Big), 
\end{equation}
where $\hat{S}^{(t)}$, $\hat{I}^{(t)}$, $\hat{R}^{(t)}$ represent the predictions of the respective compartments by the neural network at step $t$.  
Further, to infer the transmission rate and recovery rate, we solve the inverse problem and optimize these parameters alongside the neural network predictions. 
We constrain them within [-1,1] using a hyperbolic tangent regularization~\cite{dubey2022activationfunctionsdeeplearning}, i.e., $\tilde{\beta} = \tanh(\beta)$ and $\tilde{\alpha} = \tanh(\alpha)$. 

Further, to enforce SIR conformity, we minimize the residuals of the governing ODEs, given in the \cref{eq:sir}.
Hence, in vector form $\mathcal{L}_\text{physics}$ (\cref{eq:loss-physics}) becomes
\begin{equation}
\mathcal{L}_{\text{physics}} = \left\|  \frac{d\hat{S}}{dt} + \tilde{\beta} \frac{\hat{S} \hat{I}}{N} \right\|^2  + \left\| \frac{d\hat{I}}{dt} - \tilde{\beta} \frac{\hat{S} \hat{I}}{N} + \tilde{\alpha} \hat{I} \right\|^2  + \left\| \frac{d\hat{R}}{dt} + \tilde{\alpha} \hat{I} \right\|^2. 
\end{equation}

\noindent Then, the total loss combines data fidelity and physics-based regularization following the standard PINN formulation, i.e., $\mathcal{L}_\text{SIR} = \mathcal{L}_\text{data} + \mathcal{L}_{\text{physics}}$.

\subsection{Time-Dependent Reproduction Number (\Rt) Estimation}
\label{sec:method2}
Estimating \Rt involves using the ODE specified in \cref{eq:rt-ode}. Following ~\cite{Millevoi2023}, 
the system consists of two state variables, $I$ and $\Rt$, with the PINN receiving $t$ as input and predicting ($\hat{I}^{(t)}$, $\Rt$). 
To approximate $I$, the PINN minimizes the squared error $|\hat{I}^{(t)} - I^{(t)}|^2$, for each $t \in \{1, \dots, T\}$. 
Hence, the corresponding data loss function $\mathcal{L}_\text{data}$ (\cref{eq:loss-data}) is:
\begin{equation}\label{eq:rsir-data}
\mathcal{L}_{data} = \frac{1}{T} \sum_{t=1}^{T} \Big|\hat{I}^{(t)} - I^{(t)}\Big|^2.
\end{equation}
Note that while the prediction $\hat{I}$ is optimized by minimizing this error, $\Rt$ is inferred solely from the residuals of the governing ODE. 
In our computation, we use the standardized scaling provided in \cite{Millevoi2023}. 
Then, the corresponding physics loss $\mathcal{L}_\text{physics}$ for the time-dependent reduced SIR model is the squared residual:
\begin{equation}
\mathcal{L}_{physics} =\bigg\| \frac{d\hat{I}}{dt} - \alpha (t_f - t_0) (\Rt - 1) \hat{I} \bigg\|^2.
\end{equation}
Training proceeds in two stages: (1), the PINN is optimized using only $\mathcal{L}_{\text{data}}$ (\cref{eq:rsir-data}) to fit the observational data.
(2), the complete loss function, defined as $\mathcal{L}_\text{rSIR} = w_0\mathcal{L}_{\text{data}} + w_1\mathcal{L}_\text{physics}$ with balancing parameters $w_0$ and $w_1$, is minimized, ensuring both data consistency and conformity with our assumed model.
\section{Data Collection and Experimental Setup}
\label{sec:setup}

We use public Robert Koch Institute (RKI) infection data~\cite{RKI,RKIa}, preprocessing raw infections~\cite{GHInf} per federal state and German death cases~\cite{GHDead} separately. 
Lacking explicit recovery data, we model a recovery queue to transition infected individuals to the removed group, aligning with typical recovery periods noted by WHO~\cite{WHO}.
We use state population sizes from 2020~\cite{EuroStat}, and the initial number of infectious individuals is taken from the original cases recorded on March 9, 2020. 
Our analysis spans March 9, 2020, to June 22, 2023 (1200 days), covering the most active phases of the COVID-19 pandemic~\cite{RKI,RKIa}.

To estimate pandemic parameters for each German state, we employ Physics-Informed Neural Networks (PINNs) to fit SIR models to observed case data, as detailed in \cref{sec:method}.
Crucially, before applying this framework to the extended RKI data, we validate its fundamental capabilities. 
Thus, we first replicate an experiment from~\cite{baerwolff2021modeling}, where Germany's early pandemic dynamics were analyzed using a classical dampened Gauss-Newton method.
Using this approach, they approximate the transmission rate $\beta$ as $0.22658$, whereas our PINN-based approach yields a consistent result of $0.22822$ on their data.
This replication confirms that \textbf{our PINN-based approach reproduces results obtained by established PDE solvers}, providing confidence in its application to more complex scenarios.
We include the full details of this validation in the supplementary material.

With our methodology validated, the core contribution of our work is the fine-grained, spatiotemporal analysis of COVID-19 dynamics across all German federal states over a three-year period.
Our focus is thus not on comparative benchmarking with other solvers, which often operate on synthetic or short-term data. 
Instead, we aim to demonstrate the practical utility of PINNs for extracting detailed insights from extensive, real-world epidemiological data.
To ensure the robustness of our findings, each experiment is repeated ten times per state.
Further, we investigate the pandemic's evolution under two separate paradigms:

\subsubsection{Time-Independent Parameter Identification:} 
\label{sec:alpha_beta_exp}
First, we estimate the transmission ($\beta$) and recovery ($\alpha$) rates for the entire pandemic by optimizing a PINN to fit the SIR model and the observed infection data, as detailed in \cref{sec:method1}. 
Here $\alpha$ and $\beta$ are trainable variables initialized within the PINN training process.

We employ a PINN architecture comprising seven hidden layers (each with 20 neurons) and hyperbolic tangent activations~\cite{Lecun1998}. 
Adhering to hyperparameter settings from~\cite{Shaier2021}, the model is subsequently trained for 10K iterations using a 0.001 learning rate with a polynomial scheduler~\cite{Paszke2019}.
The complete training process is repeated ten times for each German state to obtain state-wise estimates of $\alpha$ and $\beta$.

\subsubsection{Time-Dependent Reproduction Number (\Rt) Estimation:}
\label{sec:rt_exp}
Following \cite{Millevoi2023}, we estimate the time-dependent reproduction number ($\Rt$), assuming a constant recovery rate of $\alpha = \nicefrac{1}{14}$ for normal conditions, as noted by WHO~\cite{WHO}.
Additionally, we conduct a second experiment using the state-wise $\alpha_\text{exp}$ values determined in our first time-independent experiment.

We use the same PINN architecture, employing ReLU activations~\cite{nair2010rectified}.
To ensure the model effectively learns the infection compartment, we first optimize only the data loss ($\mathcal{L}_{\text{data}}$) for 30K iterations. 
Next, we train using the combined loss $\mathcal{L}_{\text{rSIR}}$ for 20K iterations, initializing $\Rt$ from the time-independent experiment results.
In $\mathcal{L}_{\text{rSIR}}$, we find balancing $\mathcal{L}_\text{data}$ and $\mathcal{L}_\text{physics}$ to be crucial for ensuring convergence given the different magnitudes of the loss terms.
Specifically, we multiply the data loss by $w_0=10^2$ and scale the physics term by $w_1=1\times 10^{-6}$.
For the federal states, we observe improvements when increasing both weights to $10^3$ and $4\times10^{-6}$ respectively, due to the smaller population sizes.
As before, each experiment is repeated ten times to ensure robustness.

\begin{figure}[t]
    \centering
    \includegraphics[width=\textwidth]{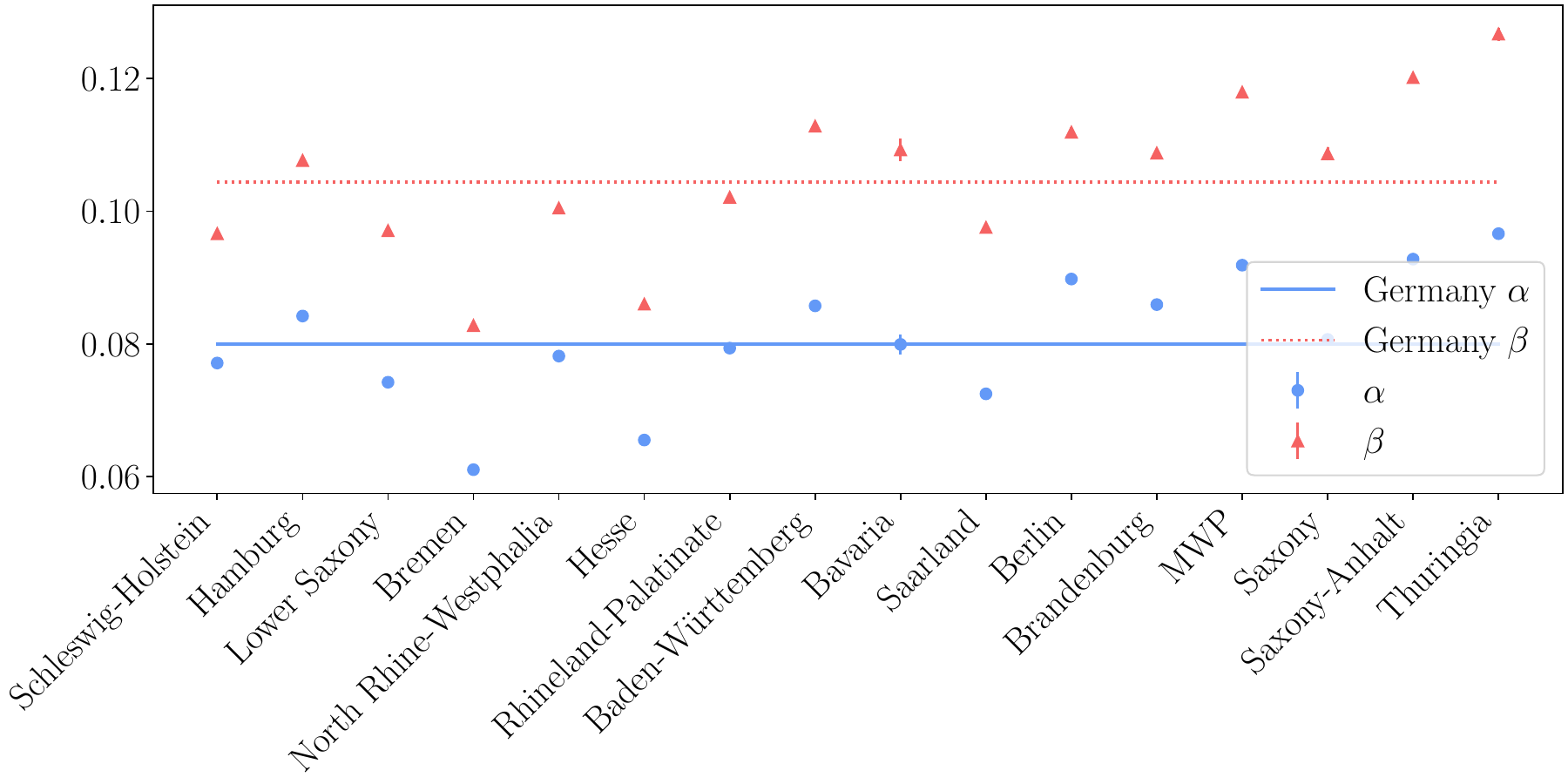}
    \caption{Visualization of the recovery rate $\alpha$ and the transmission rate $\beta$ for each federal state (MWP=Mecklenburg-Western Pomerania) compared to the mean values of $\alpha$ and $\beta$ for Germany.}
    \label{fig:alpha_beta_mean_std}
\end{figure}

\begin{table}[t]
    \begin{center}
        \caption{Pandemic parameter means and standard deviations for Germany, and each German state (MWP=Mecklenburg-Western Pomerania, NRW=North Rhine-Westphalia). 
        Furthermore, we include the vaccination percentage provided by the German Fed. Ministry for Health~\cite{FMH} and corresponding population sizes~\cite{EuroStat}. 
        }
        \setlength{\tabcolsep}{7pt}
        \label{table:state_mean_std}
        \begin{tabular}{lrccccc}
            \toprule
            State                & $N$ [$10^6$] & $\alpha$                   & $\beta$                    & Vaccinations [\%] \\
            \midrule
            Germany              & 83.16        & 0.080 {\tiny $\pm 0.000 $} & 0.104 {\tiny $\pm 0.001 $} & 76.4 \\ 
            \midrule
            Schleswig-Holstein   & 2.90         & 0.077 {\tiny $\pm 0.000 $} & 0.097 {\tiny $\pm 0.000 $} & 79.5 \\ 
            Hamburg              & 1.84         & 0.084 {\tiny $\pm 0.000 $} & 0.108 {\tiny $\pm 0.001 $} & 84.5 \\ 
            Lower Saxony         & 7.99         & 0.074 {\tiny $\pm 0.001 $} & 0.097 {\tiny $\pm 0.001 $} & 77.6 \\ 
            Bremen               & 0.68         & 0.061 {\tiny $\pm 0.000 $} & 0.083 {\tiny $\pm 0.000 $} & 88.3 \\ 
            NRW                  & 17.94        & 0.078 {\tiny $\pm 0.000 $} & 0.100 {\tiny $\pm 0.001 $} & 79.5 \\ 
            Hesse                & 6.29         & 0.066 {\tiny $\pm 0.001 $} & 0.086 {\tiny $\pm 0.001 $} & 75.8 \\ 
            Rhineland-Palatinate & 4.08         & 0.079 {\tiny $\pm 0.001 $} & 0.102 {\tiny $\pm 0.001 $} & 75.6 \\ 
            Baden-Württemberg    & 11.07        & 0.086 {\tiny $\pm 0.000 $} & 0.113 {\tiny $\pm 0.001 $} & 74.5 \\
            Bavaria              & 13.10        & 0.080 {\tiny $\pm 0.001 $} & 0.109 {\tiny $\pm 0.002 $} & 75.1 \\ 
            Saarland             & 0.99         & 0.072 {\tiny $\pm 0.000 $} & 0.098 {\tiny $\pm 0.001 $} & 82.4 \\ 
            Berlin               & 3.67         & 0.090 {\tiny $\pm 0.001 $} & 0.112 {\tiny $\pm 0.001 $} & 78.1 \\ 
            Brandenburg          & 2.52         & 0.086 {\tiny $\pm 0.001 $} & 0.109 {\tiny $\pm 0.001 $} & 68.1 \\ 
            MWP                  & 1.61         & 0.092 {\tiny $\pm 0.000 $} & 0.118 {\tiny $\pm 0.000 $} & 74.7 \\ 
            Saxony               & 4.07         & 0.081 {\tiny $\pm 0.001 $} & 0.109 {\tiny $\pm 0.001 $} & 65.1 \\ 
            Saxony-Anhalt        & 2.20         & 0.093 {\tiny $\pm 0.000 $} & 0.120 {\tiny $\pm 0.000 $} & 74.1 \\ 
            Thuringia            & 2.13         & 0.097 {\tiny $\pm 0.001 $} & 0.127 {\tiny $\pm 0.001 $} & 70.3 \\
            \bottomrule
        \end{tabular}
    \end{center}
\end{table}
\section{Results and Discussion}

Our main contribution is a spatiotemporal analysis of the COVID-19 pandemic at the level of individual German states over a 1,200-day period, using a PINN framework. 
This analysis enables us to identify local differences and long-term trends in the pandemic’s progression that are not apparent in national-level summaries. 
First, we discuss overall transmission and recovery rates for each state. 
Next, we detail state-wise time-dependent reproduction numbers \Rt.
\begin{table}[t]
     \begin{center}
         \caption{
         Average number of days with $\Rt > 1$, and the average peak \Rt values for all German states (MWP=Mecklenburg-Western Pomerania, NRW=North Rhine-Westphalia) and Germany, for $\alpha=\nicefrac{1}{14}$ and $\alpha_{\text{exp}}$ (see \cref{table:state_mean_std}).
         }
         \label{table:state_error}
         \setlength{\tabcolsep}{10pt}
         \begin{tabular}{lcccccc}
             \toprule
                     & \multicolumn{2}{c}{days with $\Rt>1$} & \phantom{0} & \multicolumn{2}{c}{peak $\Rt$}                                                                       \\
             \cmidrule{2-3}\cmidrule{5-6}
             State name           & $\alpha=\nicefrac{1}{14}$ & $\alpha_{\text{exp}}$ & \phantom{0} & $\alpha=\nicefrac{1}{14}$ & $\alpha_{\text{exp}}$ \\
             \midrule
             Germany              & 312.0                 & 301.5                & \phantom{0} & 1.643                 & 1.705                  \\ 
             \midrule
             Schleswig-Holstein   & 352.1                 & 355.6                & \phantom{0} & 1.525                 & 1.441                  \\ 
             Hamburg              & 398.3                 & 316.0                & \phantom{0} & 1.689                 & 1.577                  \\ 
             Lower Saxony         & 327.9                 & 298.4                & \phantom{0} & 1.637                 & 1.682                  \\ 
             Bremen               & 326.1                 & 402.9                & \phantom{0} & 1.508                 & 1.525                  \\ 
             NRW                  & 280.1                 & 316.8                & \phantom{0} & 1.954                 & 1.789                  \\ 
             Hesse                & 344.1                 & 308.8                & \phantom{0} & 1.774                 & 1.750                  \\ 
             Rhineland-Palatinate & 341.7                 & 335.5                & \phantom{0} & 1.582                 & 1.515                  \\ 
             Baden-Württemberg    & 372.2                 & 307.0                & \phantom{0} & 1.617                 & 1.608                  \\ 
             Bavaria              & 342.9                 & 321.2                & \phantom{0} & 1.719                 & 1.532                  \\ 
             Saarland             & 388.1                 & 338.9                & \phantom{0} & 1.495                 & 1.547                  \\ 
             Berlin               & 304.7                 & 305.7                & \phantom{0} & 1.686                 & 1.485                  \\ 
             Brandenburg          & 380.2                 & 376.6                & \phantom{0} & 1.795                 & 1.466                  \\ 
             MWP                  & 399.8                 & 327.9                & \phantom{0} & 1.645                 & 1.375                  \\ 
             Saxony               & 368.1                 & 368.9                & \phantom{0} & 1.696                 & 1.523                  \\ 
             Saxony-Anhalt        & 345.8                 & 335.9                & \phantom{0} & 1.706                 & 1.424                  \\ 
             Thuringia            & 373.7                 & 387.2                & \phantom{0} & 1.959                 & 1.429                  \\
             
             \bottomrule
         \end{tabular}
     \end{center}
\end{table}

\begin{figure}[t]

    \centering
        \includegraphics[width=\textwidth]{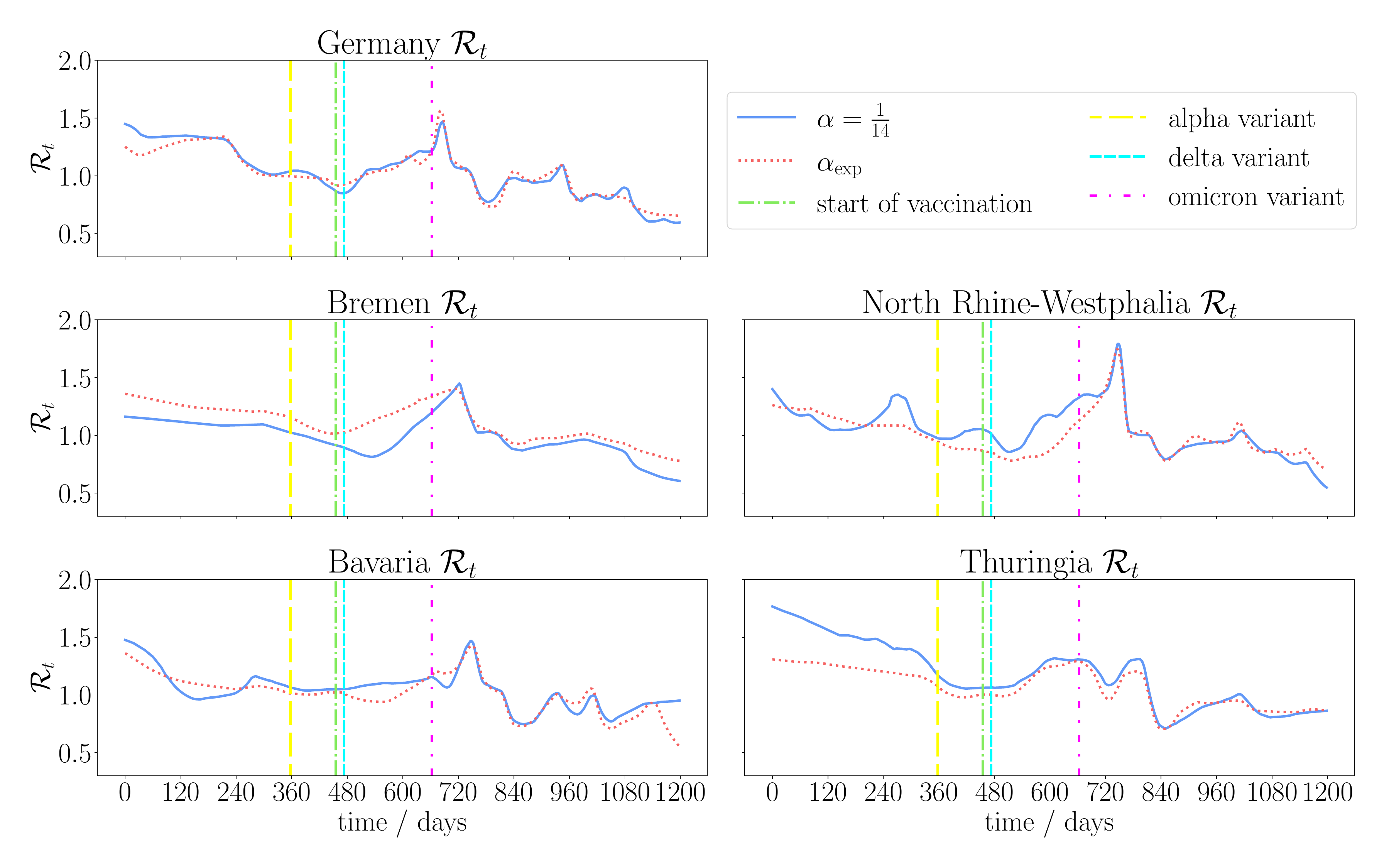}
    \caption{
    Visualization of the time-dependent reproduction number $\Rt$ over the pandemic for Germany on top, followed by various states.
    In all cases, we show results for $\alpha = \nicefrac{1}{14}$ \cite{WHO} and our experimentally determined state-specific recovery rate $\alpha_{\text{exp}}$.
    Events~\cite{COVIDChronik} like the peak of specific virus variants or the start of the vaccination campaigns are marked horizontally. 
    The remaining states are included in the supplementary material.
    }
    \label{fig:r_t_evolution}
\end{figure}

\subsubsection{Time-Independent Parameter Identification:} 
\cref{table:state_mean_std} presents results for our first setup, showing the estimated recovery rate $\alpha$ and transmission rate $\beta$ for each German state, along with vaccination percentages as reported by the German Federal Ministry for Health~\cite{FMH}. 
The results highlight significant regional variations in pandemic dynamics. \cref{fig:alpha_beta_mean_std} further visualizes the state-wise parameters relative to the national values.

A key observation is that Saxony-Anhalt and Thuringia exhibit the highest transmission rates of 0.120 and 0.127, respectively. 
In contrast, Bremen and Hesse have the lowest transmission rates with 0.083 and 0.086.
Crucially, these differences align with regional vaccination rates.
For instance, Bremen, with the highest vaccination percentage (88.3\%), also has the lowest $\beta$.
Notably, we find a significant negative correlation of $-0.5741$ ($p=0.02$) between state-level transmission rates and vaccination percentages, reinforcing the link between higher vaccination coverage and reduced disease spread (see \cref{fig:beta_correlation}).

The recovery rates follow a comparable trend, with regions experiencing higher $\beta$ also showing higher $\alpha$. 
Thuringia (0.097), Saxony-Anhalt (0.093), and MWP (0.092) all exhibit above-average recovery rates, while Bremen (0.061) and Hesse (0.066) show lower recovery rates. 
This suggests that the turnover of cases is faster in areas with higher infection rates, possibly due to a combination of natural immunity effects and healthcare responses.
Generally, the recovery rates remain close to the assumed 14-day recovery period ($\alpha \approx \nicefrac{1}{14} = 0.0714$), with most states falling within a reasonable distance.

While our approach generally captures typical COVID-19 recovery dynamics across regions, certain outliers warrant discussion. 
Saxony, for instance, shows a relatively low transmission rate despite the lowest vaccination rate (65.1\%), potentially due to a sparse population density or underreported case numbers.
Conversely, Berlin exhibits an above-average $\beta$ of 0.112 despite a high vaccination percentage (78.1\%). 
We hypothesize that Berlin's high population density and population mobility, fostering increased social interactions, are a potential reason offsetting some of the vaccination benefits.

Geographically, these findings highlight broader regional pandemic differences across Germany.
Eastern states like Thuringia and Saxony-Anhalt consistently show higher transmission and recovery rates, likely influenced by lower vaccination coverage. 
In contrast, western and northern states such as Saarland, Hesse, Bremen, and Schleswig-Holstein tend towards lower transmission, potentially benefiting from higher vaccine uptake.
To further investigate the different transmission dynamics, we estimate the time-dependent reproduction number \Rt.
In the following, $\alpha_{\text{exp}}$ represents the recovery rates from \cref{table:state_mean_std}.

\subsubsection{Time-Dependent Reproduction Number \Rt Estimation:}
\cref{table:state_error} summarizes \Rt estimates for all 16 federal states and Germany as a whole.
In particular, we provide the number of days with $\Rt>1$ and the peak reproduction number reached during the pandemic. 
Specifically, these results facilitate insights into the severity of outbreaks (peak $\Rt$ values) and the duration of active transmission (days with $\Rt > 1$). 
Further, \cref{fig:r_t_evolution} visualizes the \Rt during the pandemic for selected states (see supplementary for the remaining).
Major pandemic events, including the start of vaccinations and the emergence of the Alpha, Delta, and Omicron variants, are annotated in \cref{fig:r_t_evolution} to contextualize observations.

We find a strong connection between transmission rates and the duration of the pandemic in each region. 
Thuringia, which exhibited the highest estimated transmission rate $\beta$, also experienced the longest period where $\Rt > 1$, lasting nearly 387 days under the state-specific recovery rate $\alpha_{\text{exp}}$. 
Saxony-Anhalt and Brandenburg show similar trends, reinforcing their previously observed high transmission rates.
Peak $\Rt$ values further reflect transmission intensity. 
The highest peak is observed in Thuringia (1.959) under the fixed recovery rate assumption ($\alpha = \nicefrac{1}{14}$), closely matching the ranking of the time-independent analysis. 
Similarly, Saxony-Anhalt and Brandenburg, which had high transmission rates, also show elevated peak $\Rt$ values. 
In contrast, Bremen, Schleswig-Holstein, and Saarland display lower peaks, correlating with their lower transmission rates. 
Similar to our first experiment, we find a negative correlation (-0.44552, $p=0.079$) between peak $\Rt$ and the regional vaccination rates, further reinforcing the effectiveness of vaccination (see \cref{fig:beta_correlation}).

Regarding our setup, we find that using state-specific recovery rates ($\alpha_{\text{exp}}$) generally reduces peak $\Rt$ values across most states, highlighting the role of localized recovery dynamics in modeling pandemic progression. 
For example, Bavaria’s peak $\Rt$ declines from 1.719 to 1.532, while Brandenburg’s drops from 1.795 to 1.466. 
However, some states, such as Thuringia and Bremen, experience a longer period where $\Rt > 1$, suggesting that variation in recovery rates influences how long infections persist.
Overall, we observe similar rankings for both settings, further validating our PINN-based framework for real-world epidemiological data.

Next, we examine the temporal evolution of $\Rt$ in \cref{fig:r_t_evolution} to contextualize our findings. 
At the onset of the pandemic, $\Rt$ exceeded $1.2$ in multiple states, signaling rapid early transmissions before lockdowns and social distancing measures successfully reduced it below 1. 
The start of vaccinations led to a gradual decline in $\Rt$, particularly in highly vaccinated states like Bremen and Schleswig-Holstein. 
However, the effect was delayed as immunity built up over time.
The Omicron variant, emerging in late 2021, led to the highest recorded peaks of $\Rt$, reflecting its increased transmissibility despite widespread vaccination efforts.
Even highly vaccinated states like Bremen experienced short-lived but notable peaks, underscoring the severity of the Omicron variant.

Following the Omicron wave, transmission rates steadily declined across all states. By mid-2022, $\Rt$ stabilized near or below 1, though regional variations remained. 
The rate of decline differed between states, likely due to booster uptake, regional mitigation strategies, and behavioral factors~\cite{RKI,RKIa,WHO}.
\section{Conclusions}

In this work, we model the COVID-19 dynamics in Germany.
Specifically, our contributions are state-wise analyses over long time scales.
To perform this spatio-temporal analysis, we use Physics Informed Neural Networks (PINNs) \cite{Raissi2019} to solve the inverse problem of inferring pandemic parameters from observational data.
We conducted our study on data recorded by the Robert Koch Institute (RKI) between March 9, 2020, and June 22, 2023, encompassing 1200 days of the pandemic.
By solving the inverse problem for the standard SIR model \cite{Kermack1927}, we estimate the transmission rates $\beta$ and the recovery rates $\alpha$ per federal state.
Afterward, we approximate the time-dependent reproduction number \Rt.

We find strong regional variations: states like Saxony-Anhalt and Thuringia exhibited the highest transmission rates $\beta$, while Bremen and Hesse had the lowest. 
Further, a strong negative correlation between vaccination rates and transmission rates $\beta$ and peak reproduction number \Rt indicates vaccination’s role in reducing transmission.
Regarding the estimation of \Rt, the model effectively captured the Omicron wave and its corresponding peak.
Again, regional factors strongly influenced outcomes: Thuringia’s low vaccination rate coincides with its high \Rt peak, while Berlin’s dense population likely amplified transmission despite higher vaccination numbers.
Overall, our findings emphasize that regional heterogeneity strongly influenced the pandemic's local development in Germany. 
We demonstrate the utility of PINNs in epidemiological modeling, offering a data-driven framework to solve inverse problems for analyzing pandemic dynamics at a sub-national level.
By integrating physics-based disease models with real-world observational data, PINNs provide a powerful approach to understanding infectious disease spread.

\noindent\textbf{Future Work:}
While our focus on comparably simple SIR models enables the analysis of abundantly available standard epidemiological data on the regional level, our work is limited by the corresponding assumptions.
For instance, we do not account for factors such as age-stratified transmission, reinfections, or mobility-driven spread.
Hence, future work should explore more complex epidemiological models (e.g., SVIHR~\cite{Shaier2021}, SEIR~\cite{rahman2021review}, SAIRD~\cite{han2024approaching}, agent-based simulations~\cite{Kerr2021}) in a local setup or incorporate additional data sources such as contact tracing, seasonal effects, and behavioral interventions~\cite{Bodine2020,Maziarz2020}.

\bibliographystyle{splncs04}
\bibliography{sources}

\begin{thebibliography}{10}
\providecommand{\url}[1]{\texttt{#1}}
\providecommand{\urlprefix}{URL }
\providecommand{\doi}[1]{https://doi.org/#1}

\bibitem{baerwolff2021modeling}
B{\"a}rwolff, G.: Modeling of covid-19 propagation with compartment models. Mathematische Semesterberichte  \textbf{68}(2),  181--219 (Oct 2021). \doi{10.1007/s00591-021-00312-9}, \url{https://doi.org/10.1007/s00591-021-00312-9}

\bibitem{Berkhahn2022}
Berkhahn, S., Ehrhardt, M.: A physics-informed neural network to model covid-19 infection and hospitalization scenarios. Advances in Continuous and Discrete Models  \textbf{2022}(1) (Oct 2022). \doi{10.1186/s13662-022-03733-5}

\bibitem{Bhatkar2023-fy}
Bhatkar, S., Ma, M., Zsolway, M., Tarafder, A., Doniach, S., Bhanot, G.: Asymmetry in the peak in covid-19 daily cases and the pandemic r-parameter. medRxiv  (2023). \doi{10.1101/2023.07.23.23292960}, \url{https://www.medrxiv.org/content/early/2023/08/01/2023.07.23.23292960}

\bibitem{Bodine2020}
Bodine, E.N., Panoff, R.M., Voit, E.O., Weisstein, A.E.: Agent-based modeling and simulation in mathematics and biology education. Bulletin of Mathematical Biology  \textbf{82}(8) (Jul 2020). \doi{10.1007/s11538-020-00778-z}

\bibitem{dubey2022activationfunctionsdeeplearning}
Dubey, S.R., Singh, S.K., Chaudhuri, B.B.: Activation functions in deep learning: A comprehensive survey and benchmark. Neurocomputing  \textbf{503},  92--108 (2022). \doi{https://doi.org/10.1016/j.neucom.2022.06.111}, \url{https://www.sciencedirect.com/science/article/pii/S0925231222008426}

\bibitem{EuroStat}
{Eurostat}: Population on 1 january by age, sex and nuts 2 region. \url{https://ec.europa.eu/eurostat/databrowser/bookmark/6b9a13dc-185a-4b23-b52a-5afae4a4e28c?lang=en}. \doi{10.2908/DEMO_R_D2JAN}, {Accessed: 2025-02-26}

\bibitem{GAGNON2023101258}
Gagnon, J.E., Kamin, S.B., Kearns, J.: The impact of the covid-19 pandemic on global gdp growth. Journal of the Japanese and International Economies  \textbf{68},  101258 (2023). \doi{https://doi.org/10.1016/j.jjie.2023.101258}, \url{https://www.sciencedirect.com/science/article/pii/S0889158323000138}

\bibitem{COVIDChronik}
{German Federal Ministry of Health}: Coronavirus-pandemie: Was geschah wann? \url{https://www.bundesgesundheitsministerium.de/coronavirus/chronik-coronavirus.html}, {Accessed: 2024-09-05}

\bibitem{FMH}
{German Federal Ministry of Health}: Übersicht zum impfstatus - covid-19-impfung in deutschland bis zum 8. april 2023. \url{https://impfdashboard.de/}, {Accessed: 2024-09-08}

\bibitem{han2024approaching}
Han, S., Stelz, L., Stoecker, H., Wang, L., Zhou, K.: Approaching epidemiological dynamics of covid-19 with physics-informed neural networks. Journal of the Franklin Institute  \textbf{361}(6),  106671 (2024). \doi{https://doi.org/10.1016/j.jfranklin.2024.106671}, \url{https://www.sciencedirect.com/science/article/pii/S0016003224000929}

\bibitem{Hethcote1989}
Hethcote, H.W.: Three Basic Epidemiological Models, pp. 119--144. Springer Berlin Heidelberg, Berlin, Heidelberg (1989). \doi{10.1007/978-3-642-61317-3_5}, \url{https://doi.org/10.1007/978-3-642-61317-3_5}

\bibitem{bookinvprob}
Isakov, V.: Inverse Problems for Partial Differential Equations. Springer Publishing Company, Incorporated, 3rd edn. (2018)

\bibitem{Kermack1927}
Kermack, W.O., McKendrick, A.G.: A contribution to the mathematical theory of epidemics. Proceedings of the Royal Society of London. Series A, Containing Papers of a Mathematical and Physical Character  \textbf{115}(772),  700--721 (Aug 1927). \doi{10.1098/rspa.1927.0118}

\bibitem{Kerr2021}
Kerr, C.C., Stuart, R.M., Mistry, D., Abeysuriya, R.G., Rosenfeld, K., Hart, G.R., Núñez, R.C., Cohen, J.A., Selvaraj, P., Hagedorn, B., George, L., Jastrzębski, M., Izzo, A.S., Fowler, G., Palmer, A., Delport, D., Scott, N., Kelly, S.L., Bennette, C.S., Wagner, B.G., Chang, S.T., Oron, A.P., Wenger, E.A., Panovska-Griffiths, J., Famulare, M., Klein, D.J.: Covasim: An agent-based model of covid-19 dynamics and interventions. PLOS Computational Biology  \textbf{17}(7),  e1009149 (Jul 2021). \doi{10.1371/journal.pcbi.1009149}

\bibitem{Kingma2014}
Kingma, D., Ba, J.: Adam: A method for stochastic optimization. International Conference on Learning Representations  (12 2014)

\bibitem{Lecun1998}
Lecun, Y., Bottou, L., Bengio, Y., Haffner, P.: Gradient-based learning applied to document recognition. Proceedings of the IEEE  \textbf{86}(11),  2278--2324 (1998). \doi{10.1109/5.726791}

\bibitem{Maziarz2020}
Maziarz, M., Zach, M.: Agent‐based modelling for sars‐cov‐2 epidemic prediction and intervention assessment: A methodological appraisal. Journal of Evaluation in Clinical Practice  \textbf{26}(5),  1352--1360 (Aug 2020). \doi{10.1111/jep.13459}

\bibitem{Millevoi2023}
Millevoi, C., Pasetto, D., Ferronato, M.: A physics-informed neural network approach for compartmental epidemiological models. PLOS Computational Biology  \textbf{20}(9),  e1012387 (2024)

\bibitem{Mwalili2020}
Mwalili, S., Kimathi, M., Ojiambo, V., Gathungu, D., Mbogo, R.: Seir model for covid-19 dynamics incorporating the environment and social distancing. BMC Res Notes  \textbf{13}(1), ~352 (2020). \doi{10.1186/s13104-020-05192-1}

\bibitem{nair2010rectified}
Nair, V., Hinton, G.E.: Rectified linear units improve restricted boltzmann machines. In: International Conference on Machine Learning (2010), \url{https://api.semanticscholar.org/CorpusID:15539264}

\bibitem{Olumoyin2021}
Olumoyin, K.D., Khaliq, A.Q.M., Furati, K.M.: Data-driven deep-learning algorithm for asymptomatic covid-19 model with varying mitigation measures and transmission rate. Epidemiologia  \textbf{2}(4),  471--489 (Sep 2021). \doi{10.3390/epidemiologia2040033}

\bibitem{Paszke2019}
Paszke, A., Gross, S., Massa, F., Lerer, A., Bradbury, J., Chanan, G., Killeen, T., Lin, Z., Gimelshein, N., Antiga, L., Desmaison, A., K\"{o}pf, A., Yang, E., DeVito, Z., Raison, M., Tejani, A., Chilamkurthy, S., Steiner, B., Fang, L., Bai, J., Chintala, S.: PyTorch: an imperative style, high-performance deep learning library. Curran Associates Inc., Red Hook, NY, USA (2019)

\bibitem{rahman2023modelling}
Rahman, A., Kuddus, M.A., Ip, R.H.L., Bewong, M.: Modelling covid-19 pandemic control strategies in metropolitan and rural health districts in new south wales, australia. Scientific Reports  \textbf{13}(1),  10352 (Jun 2023). \doi{10.1038/s41598-023-37240-8}, \url{https://doi.org/10.1038/s41598-023-37240-8}

\bibitem{rahman2021review}
Rahman, A., Kuddus, M.A., Ip, R.H., Bewong, M.: A review of covid-19 modelling strategies in three countries to develop a research framework for regional areas. Viruses  \textbf{13}(11), ~2185 (2021)

\bibitem{Raissi2019}
Raissi, M., Perdikaris, P., Karniadakis, G.: Physics-informed neural networks: A deep learning framework for solving forward and inverse problems involving nonlinear partial differential equations. Journal of Computational Physics  \textbf{378},  686--707 (Feb 2019). \doi{10.1016/j.jcp.2018.10.045}

\bibitem{RKI}
{Robert Koch Institute}: Covid-19-strategiepapiere und nationaler pandemieplan. \url{https://www.rki.de/DE/Content/InfAZ/N/Neuartiges_Coronavirus/ZS/Pandemieplan_Strategien.html}, {Accessed: 2024-09-06}

\bibitem{GHDead}
{Robert Koch Institute}: Github covid-19-todesfälle in deutschland. \url{https://github.com/robert-koch-institut/COVID-19-Todesfaelle_in_Deutschland}, {Accessed: 2024-09-05}

\bibitem{GHInf}
{Robert Koch Institute}: Github sars-cov-2 infektionen in deutschland. \url{https://github.com/robert-koch-institut/SARS-CoV-2-Infektionen_in_Deutschland}, {Accessed: 2024-09-05}

\bibitem{RKIa}
{Robert Koch Institute}: Sars-cov-2: Virologische basisdaten sowie virusvarianten im zeitraum von 2020 - 2022. \url{https://www.rki.de/DE/Content/InfAZ/N/Neuartiges_Coronavirus/Virologische_Basisdaten.html?nn=13490888#doc14716546bodyText10}, {Accessed: 2024-09-05}

\bibitem{schaback2020covid19}
Schaback, R.: On covid-19 modelling. Jahresbericht der Deutschen Mathematiker-Vereinigung  \textbf{122}(3),  167--205 (Sep 2020). \doi{10.1365/s13291-020-00219-9}, \url{https://doi.org/10.1365/s13291-020-00219-9}

\bibitem{Shaier2021}
Shaier, S., Raissi, M., Seshaiyer, P.: Data-driven approaches for predicting spread of infectious diseases through dinns: Disease informed neural networks. Letters in Biomathematics  \textbf{9}(1),  71--105 (2023). \doi{10.30707/lib9.1.1681913305.249476}

\bibitem{soilsai}
Vemuri, S.K., B{\"u}chner, T., Denzler, J.: Estimating soil hydraulic parameters for unsaturated flow using physics-informed neural networks. In: Franco, L., de~Mulatier, C., Paszynski, M., Krzhizhanovskaya, V.V., Dongarra, J.J., Sloot, P.M.A. (eds.) Computational Science -- ICCS 2024. pp. 338--351. Springer Nature Switzerland, Cham (2024). \doi{10.1007/978-3-031-63759-9_37}

\bibitem{WHO}
{World Health Organization}: Coronavirus disease (covid-19). \url{https://www.who.int/health-topics/coronavirus#tab=tab_1}, {Accessed: 2024-09-06}

\end{thebibliography}

\newpage
\appendix

\section{Reproducing Pandemic Parameters using PINNs}
\label{apx:validation}
In order to validate our method, we reproduced the results of a traditional method on real-world data. 
Bärwolff et al. ~\cite{baerwolff2021modeling} employ the damped Gauss-Newton method to find $\beta$ from a time series of data points.
Furthermore, they provide the time points, which they used to derive $\beta$ for the time span between February 13, 2020, and March 19, 2020, together with the corresponding $\beta_{\text{true}}=0.22658$. 

Just like the original study, we set $\alpha=0.07$ for the experiment. 
Our model consists of 12 hidden layers with 64 neurons each and hyperbolic tangent~\cite{Lecun1998} activation layers. 
We trained using a polynomial scheduler from PyTorch~\cite{Paszke2019}, the Adam optimizer~\cite{Kingma2014}, an initial learning rate of $1e^{-3}$, and 15K iterations. 
The observed loss is weighted by $1e^1$ in the total loss. 

As the provided data consists only of infectious data, we generated the corresponding data for the susceptible and removed compartments by utilizing~\cref{eq:sir}. 
The population size is $N=70M$ individuals, and the provided initial amount of infectious individuals is $I_0=15$.

\begin{figure}[t]
    \centering
    \begin{subfigure}{\textwidth}
        \includegraphics[width=\textwidth]{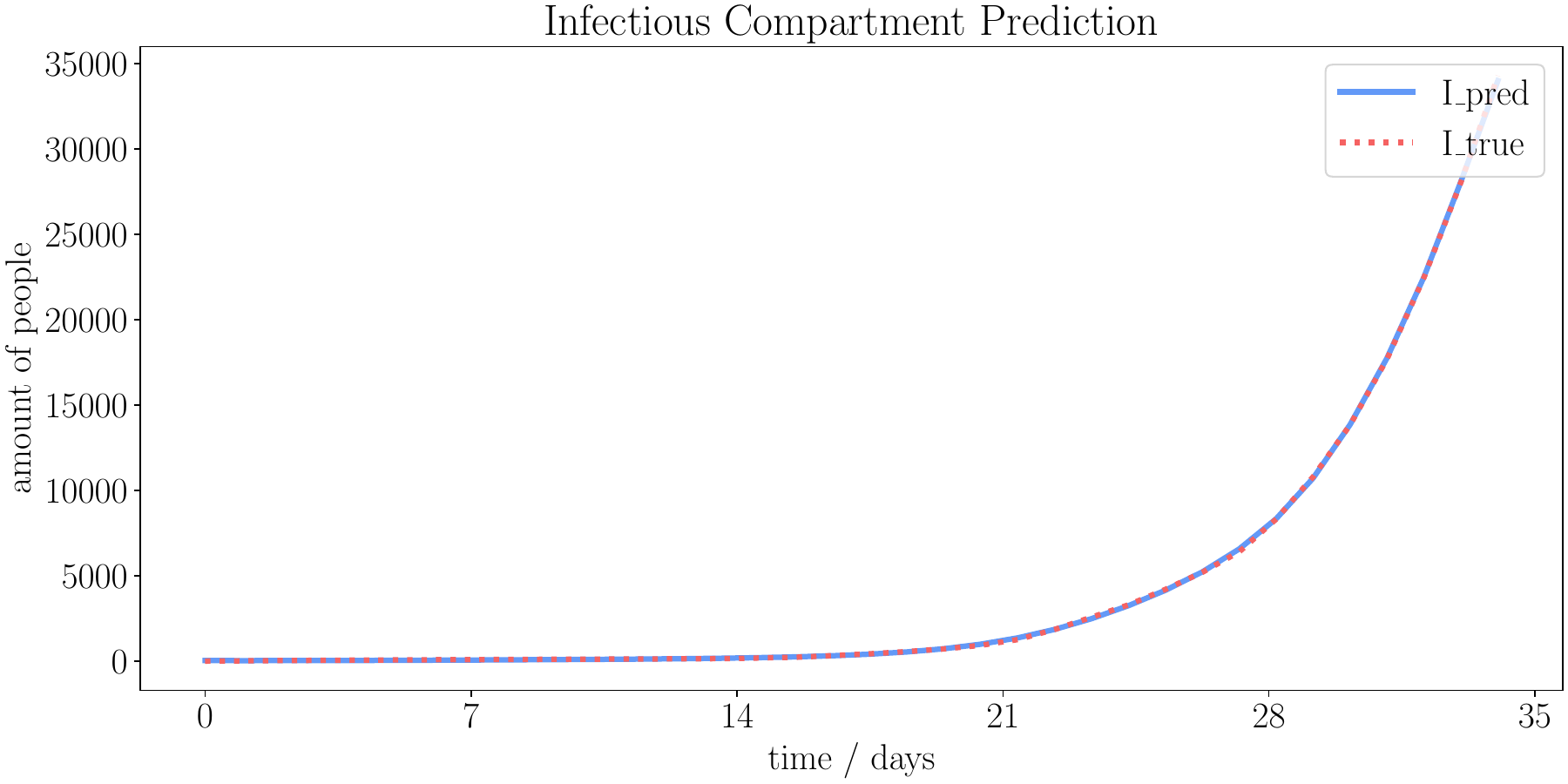}
    \end{subfigure}
    
    \caption{Visualization of the real numbers of infectious individuals in Germany and the prediction of the training with $\alpha=0.07$ for the time span between February 13, 2020, and March 19, 2020.} 
    \label{fig:paper_reproduction}
\end{figure}

In~\cref{fig:paper_reproduction}, we visualize our results from the training, demonstrating that the model successfully fits the predictions to the observed data. 
We repeated the experiment ten times, which resulted in a mean of $\beta_{\text{PINN}}=0.22822$ and a standard deviation of $\sigma^2_{\text{PINN}}=1.03367\times10^{-5}$. 
A comparison with the provided value of $\beta_{\text{true}}=0.22658$ indicates that our method has a satisfying accuracy.
Hence, we will employ PINNs to investigate longer time frames and regional variations in our main study.

\section{Additional \Rt Visualizations}
In this section, we present all results of our experiments for the estimation of the time-dependent reproduction number $\Rt$ in~\cref{sec:rt_exp}.
The analysis of \Rt across German states reveals distinct regional variations in transmission intensity and pandemic duration.
Eastern states such as Saxony, Thuringia, and Saxony-Anhalt experienced prolonged periods where \Rt > 1, aligning with their high transmission rates and lower vaccination coverage.
In contrast, northern states like Bremen, Schleswig-Holstein, and Lower Saxony exhibited lower peak values and shorter transmission durations, reflecting the effectiveness of their higher vaccination rates and public health measures. 
Southern states, including Bavaria and Baden-Württemberg, saw strong waves during Alpha and Delta but recovered faster post-Omicron, likely due to a combination of vaccine uptake and healthcare capacity. 
Western states, particularly North Rhine-Westphalia and Hesse, had moderate outbreaks but were able to manage transmission effectively, keeping \Rt under control for longer periods. 
Berlin displayed higher-than-expected peak values despite strong vaccination efforts, likely influenced by its high population density and mobility patterns, whereas Brandenburg exhibited prolonged transmission, suggesting spillover effects from Berlin. 
These findings emphasize the importance of considering regional differences in pandemic response planning, as factors such as mobility, healthcare infrastructure, and policy measures played a significant role in shaping the trajectory of COVID-19 across Germany.
\label{apx:rt}
\begin{figure}[t]
    \centering
    \begin{subfigure}{\textwidth}
        \includegraphics[width=\textwidth]{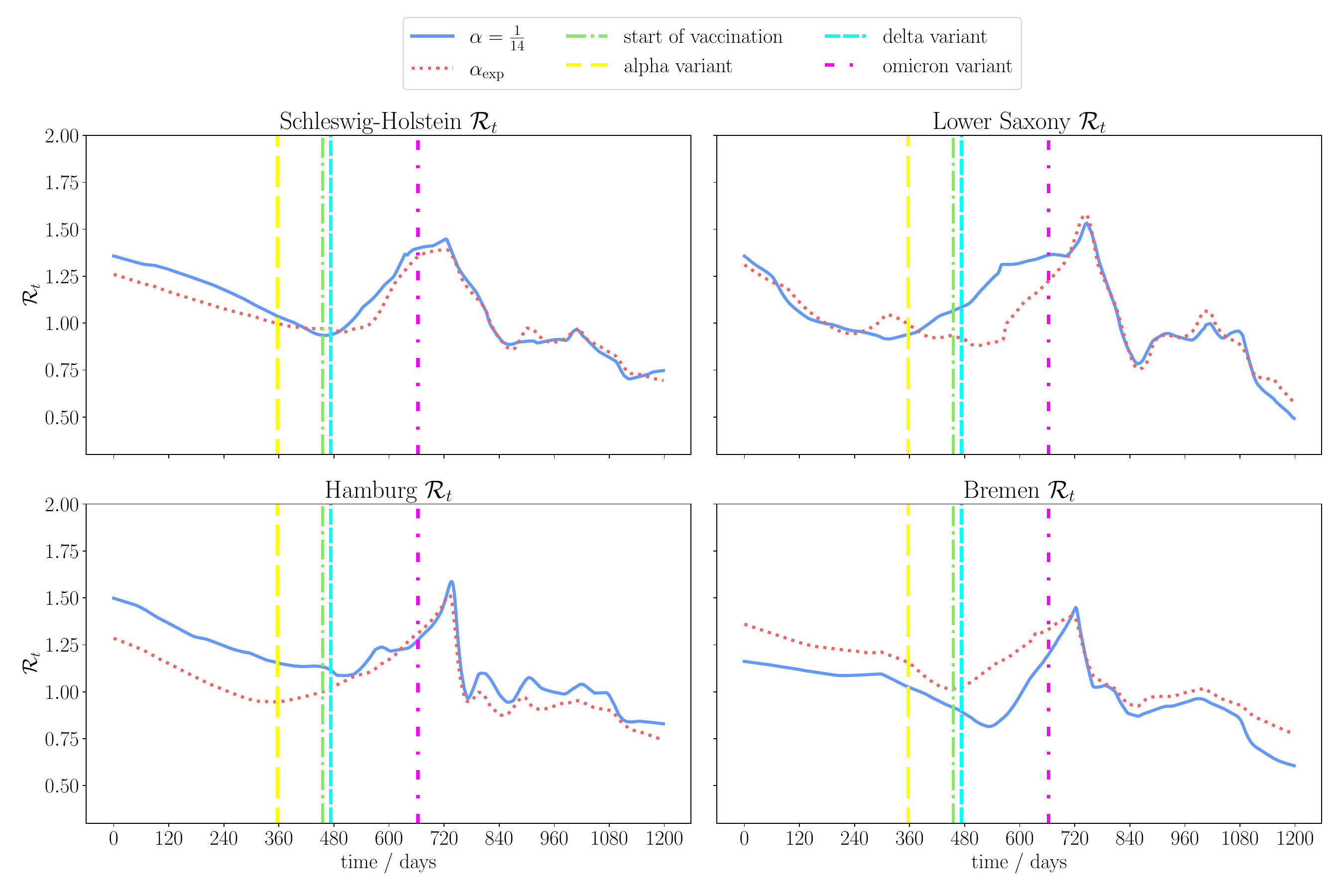}
    \end{subfigure}
    \caption{All visualizations of the $\Rt$ value from~\cref{sec:rt_exp}. (part 1)
    }
    \label{fig:state_results1}
\end{figure}
\begin{figure}[t]
    \centering
    \begin{subfigure}{\textwidth}
        \includegraphics[width=\textwidth]{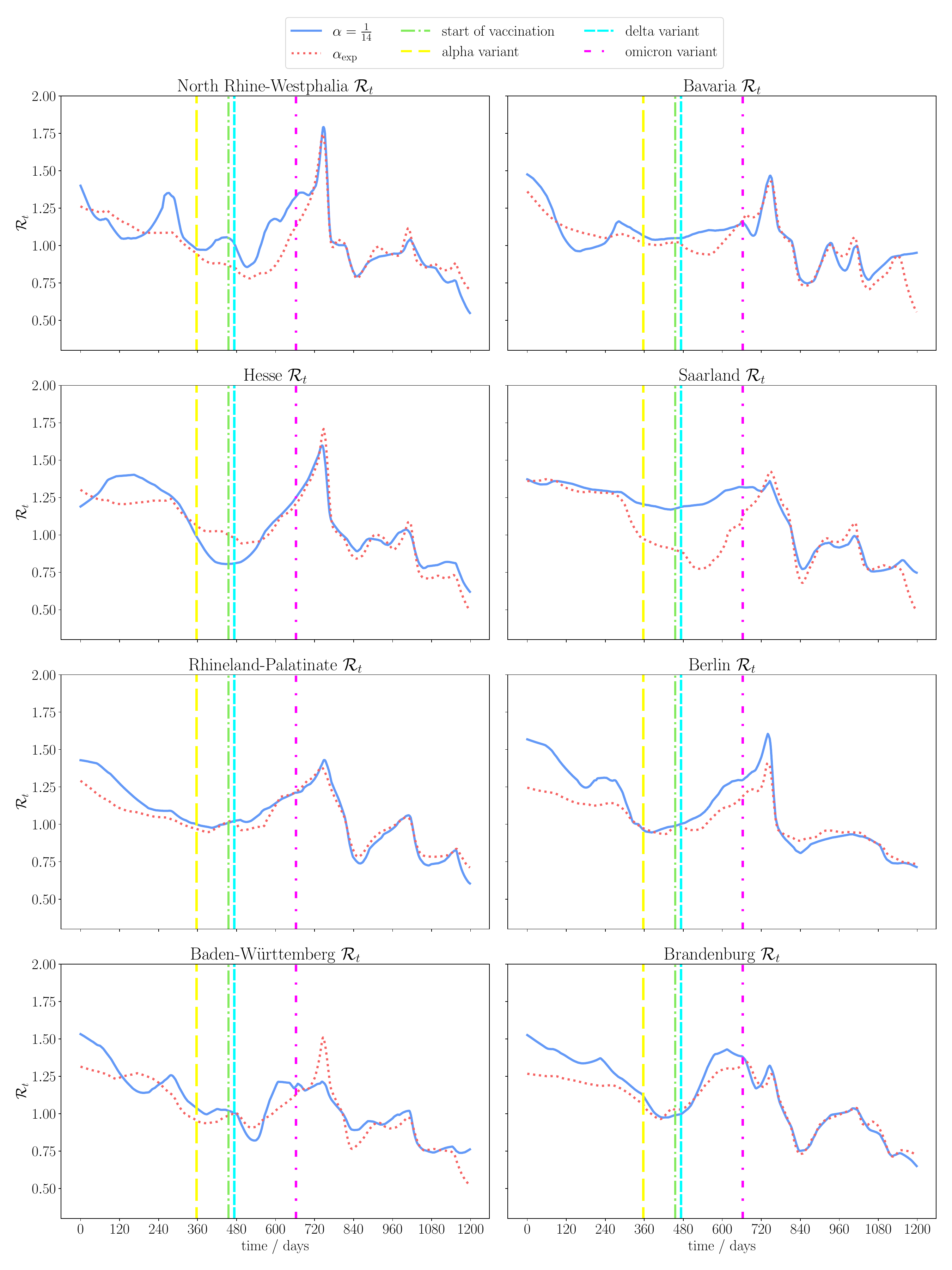}
    \end{subfigure}
    \caption{All visualizations of the $\Rt$ value from~\cref{sec:rt_exp}. (part 2)
    }
    \label{fig:state_results2}
\end{figure}
\begin{figure}[t]
    \centering
    \begin{subfigure}{\textwidth}
        \includegraphics[width=\textwidth]{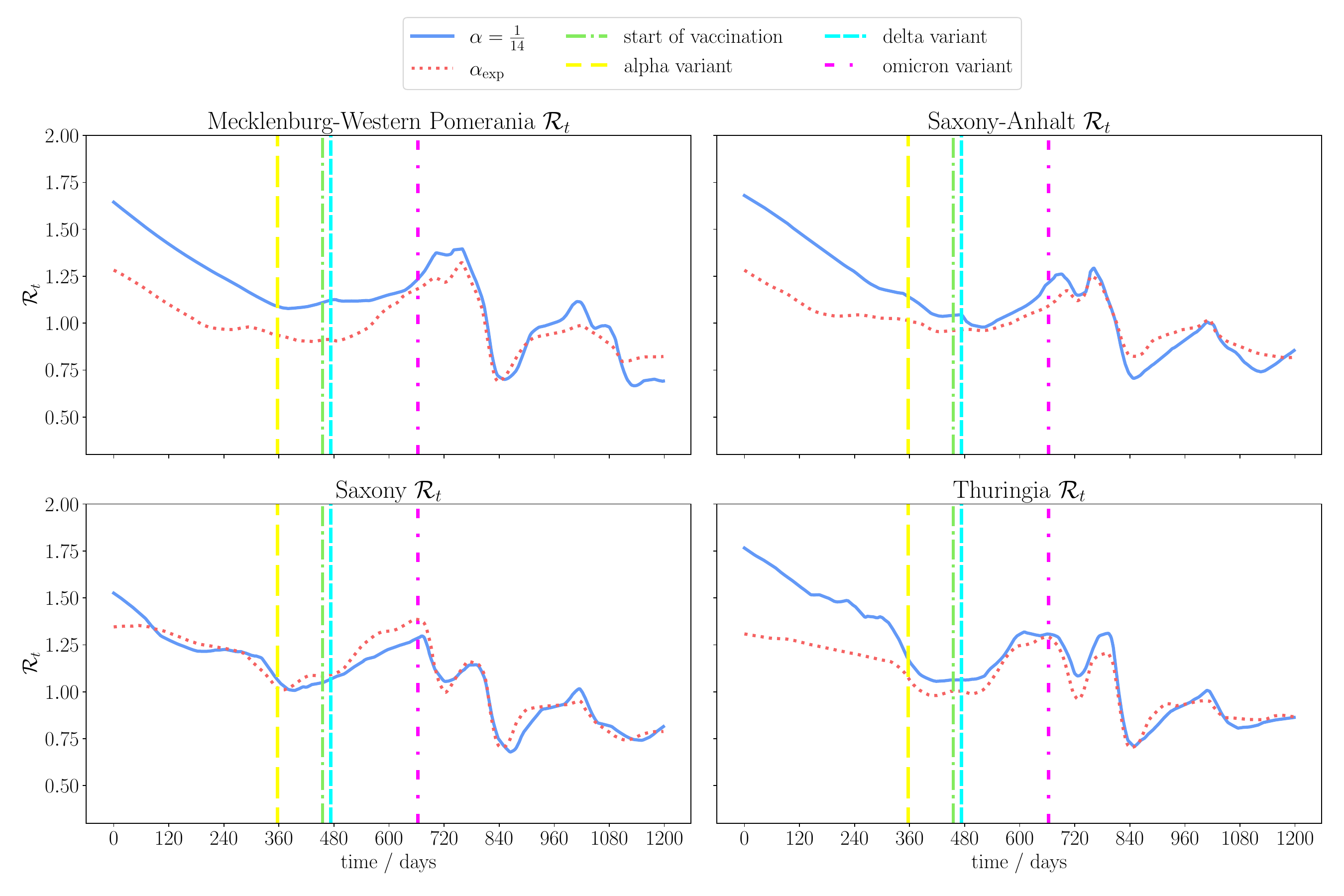}
    \end{subfigure}
    \begin{subfigure}{0.5\textwidth}
        \includegraphics[width=\textwidth]{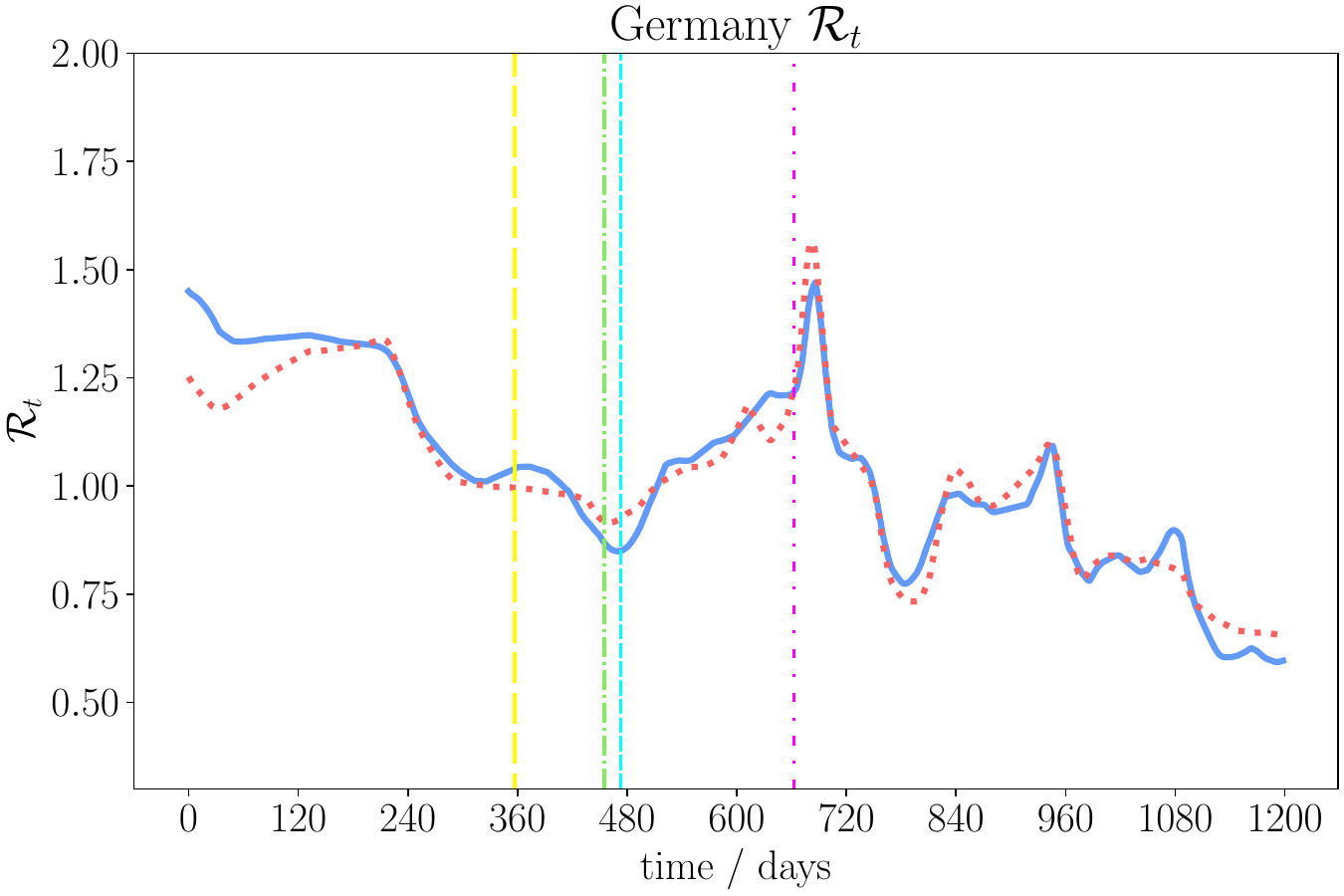}
    \end{subfigure}
    \caption{All visualizations of the $\Rt$ value from~\cref{sec:rt_exp}. (part 3)
    }
    \label{fig:state_results3}
\end{figure}

\end{document}